\documentclass[letterpaper]{article} % DO NOT CHANGE THIS
\usepackage{aaai23}  % DO NOT CHANGE THIS
\usepackage{times}  % DO NOT CHANGE THIS
\usepackage{helvet}  % DO NOT CHANGE THIS
\usepackage{courier}  % DO NOT CHANGE THIS
\usepackage[hyphens]{url}  % DO NOT CHANGE THIS
\usepackage{graphicx} % DO NOT CHANGE THIS
\usepackage{natbib}  % DO NOT CHANGE THIS AND DO NOT ADD ANY OPTIONS TO IT
\usepackage{caption} % DO NOT CHANGE THIS AND DO NOT ADD ANY OPTIONS TO IT
\usepackage{amsfonts}
\usepackage{amsmath}
\usepackage{multirow}
\usepackage{algorithm}
\usepackage{algorithmic}
\usepackage{newfloat}
\usepackage{listings}
\DeclareCaptionStyle{ruled}{labelfont=normalfont,labelsep=colon,strut=off} % DO NOT CHANGE THIS
\floatstyle{ruled}
\newfloat{listing}{tb}{lst}{}
\floatname{listing}{Listing}
\title{Efficient End-to-End Video Question Answering \\with Pyramidal Multimodal Transformer}
\author{
    Min Peng\textsuperscript{\rm 1,}\textsuperscript{\rm 2}\equalcontrib,
    Chongyang Wang\textsuperscript{\rm 3}\equalcontrib,
    Yu Shi\textsuperscript{\rm 1},
    Xiang-Dong Zhou\textsuperscript{\rm 1}
}
\affiliations{
    \textsuperscript{\rm 1}Chongqing Institute of Green and Intelligent Technology, Chinese Academy of Sciences\\
    \textsuperscript{\rm 2}University of Chinese Academy of Sciences\\
    \textsuperscript{\rm 3}Tsinghua University\\
    \{pengmin, shiyu, zhouxiangdong\}@cigit.ac.cn, mvrjustid@gmail.com
}

\begin{document}

\maketitle

\begin{abstract}
    This paper presents a new method for end-to-end Video Question Answering (VideoQA), aside from the current popularity of using large-scale pre-training with huge feature extractors. We achieve this with a pyramidal multimodal transformer (PMT) model, which simply incorporates a learnable word embedding layer, a few convolutional and transformer layers. We use the anisotropic pyramid to fulfill video-language interactions across different spatio-temporal scales. In addition to the canonical pyramid, which includes both bottom-up and top-down pathways with lateral connections, novel strategies are proposed to decompose the visual feature stream into spatial and temporal sub-streams at different scales and implement their interactions with the linguistic semantics while preserving the integrity of local and global semantics. We demonstrate better or on-par performances with high computational efficiency against state-of-the-art methods on five VideoQA benchmarks. Our ablation study shows the scalability of our model that achieves competitive results for text-to-video retrieval by leveraging feature extractors with reusable pre-trained weights, and also the effectiveness of the pyramid. Code available at: \url{https://github.com/Trunpm/PMT-AAAI23}.
\end{abstract}

\section{Introduction}

    Vision-language understanding is basic for a machine to interact with our multimodal reality. The previous success seen in computer vision and natural language processing impacted the ongoing research for a variety of vision-language understanding tasks, e.g., Visual Question Answering (VQA) \cite{REF3,REF4} and text-to-vision retrieval \cite{REF6,REF61}. Of our particular interests, Video Question Answering (VideoQA) is quite challenging, which requires accurate semantics reasoning from local-spatio regions to global-temporal dynamics of the video. This point is continuously verified in recent studies, where successes are achieved by methods that are able to capture such a multiscale property considering spatial regions of a specific frame \cite{REF47} or temporal saliency across different frames \cite{REF8,REF43}. It is time to build a VideoQA model that incorporates the learning of multiscale spatial and temporal features, preferably in an end-to-end manner.
    
\begin{figure}[t]
\centering
\includegraphics[width=\columnwidth]{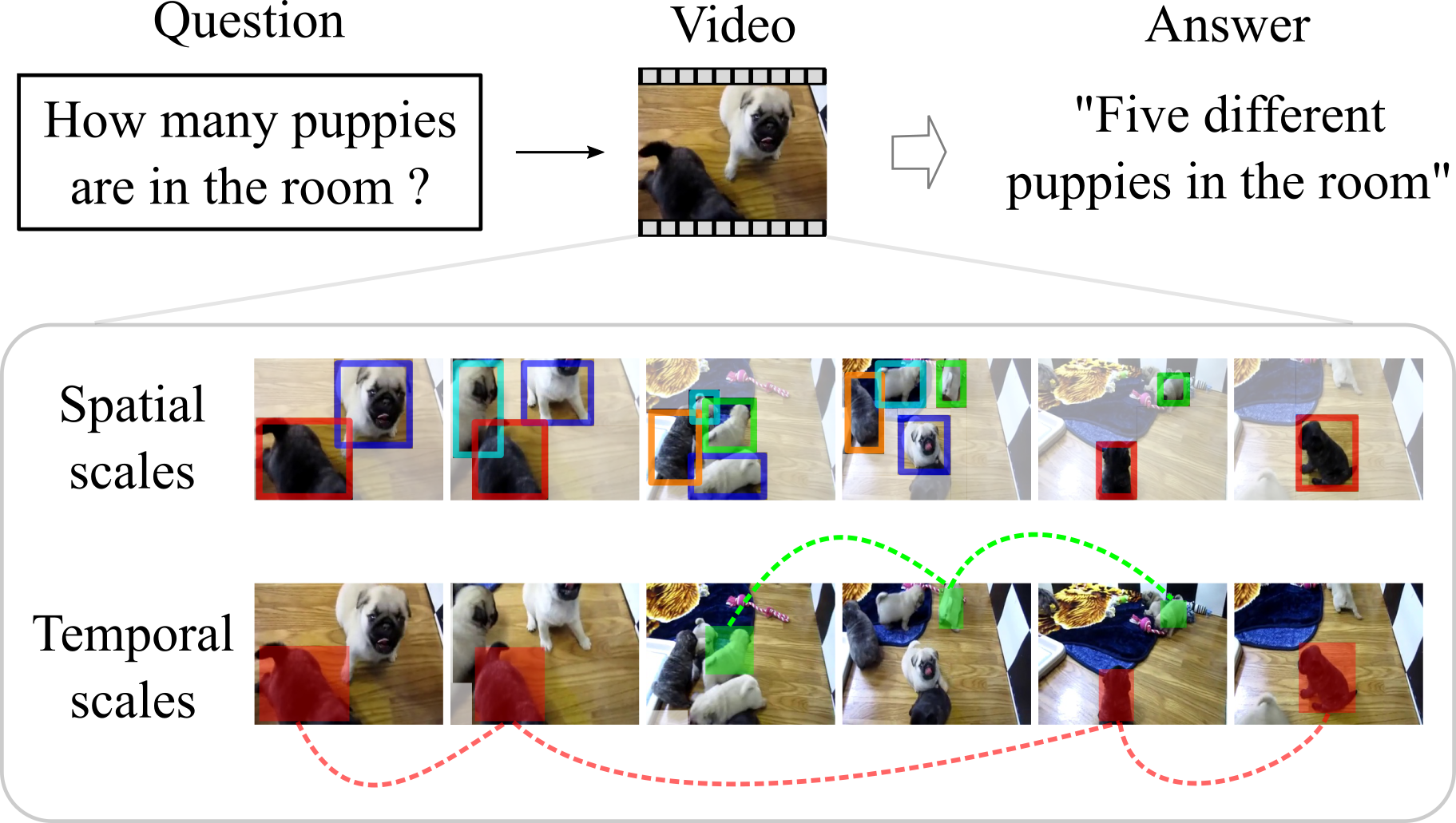}
\caption{The spatio-temporal multiscale property of VideoQA. While features of local regions (marked by anchors) contribute to the recognition of \textit{objects} per frame, their dynamics described by frames at different temporal locations (marked by anchors connected with dashed lines) help sort out their \textit{relations} or understand the \textit{events}.}
\label{fig1}
\end{figure}
    
\begin{figure*}[t]
\centering
\includegraphics[width=0.95\textwidth]{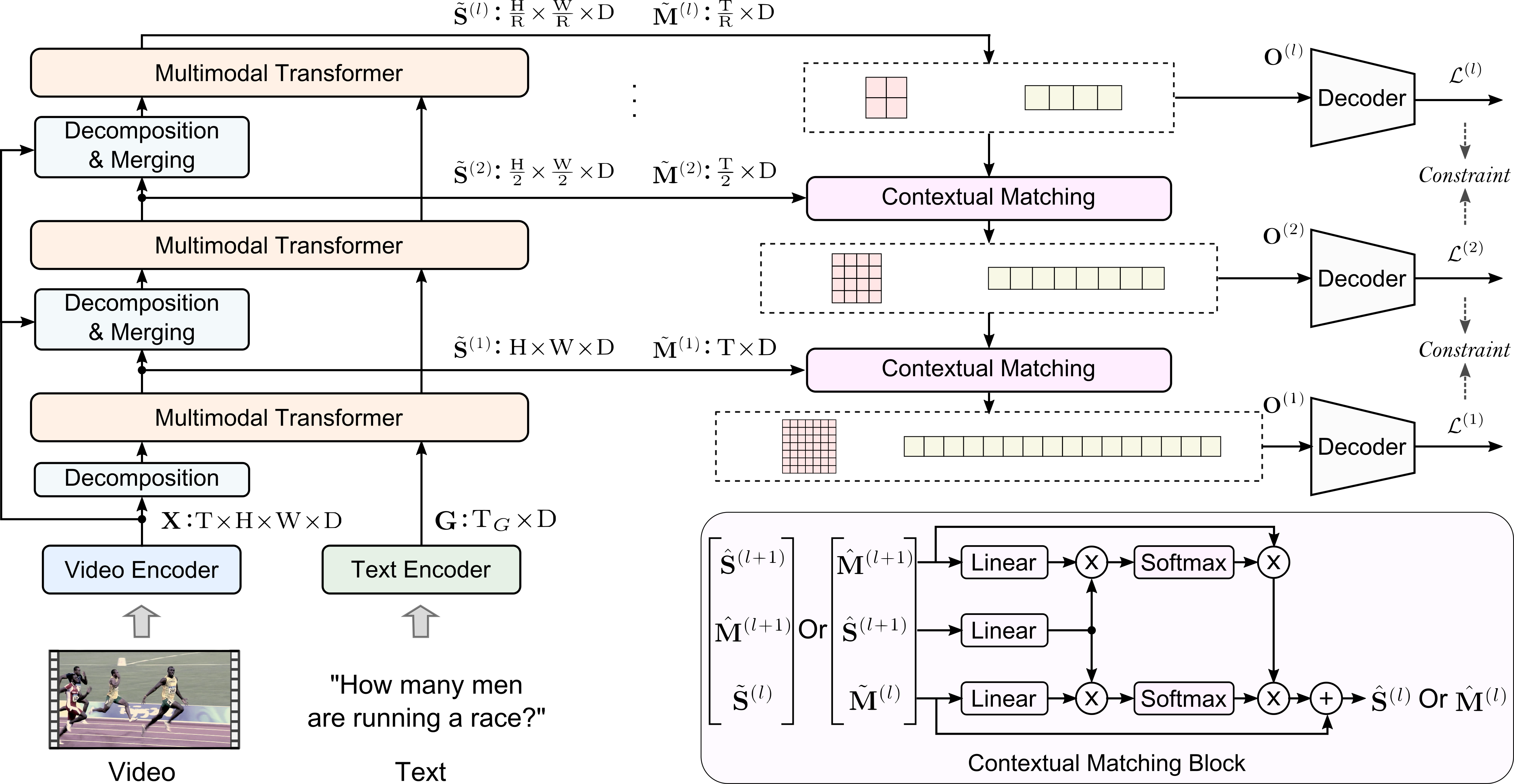}
\caption{The PMT model. Bottom-up pathway: the video and text encoders take as input raw video frames and text tokens, respectively. The visual feature $\mathbf{X}$ is decomposed into spatial $\mathbf{S}$ and temporal $\mathbf{M}$ sub-streams, and pooled to reveal their multiscale contents for multimodal learning with linguistic semantics $\mathbf{G}$. Top-down pathway: while separate losses that controlled by a constraint function are used to maintain the integrity of local and global representations, the contextual matching block therein fuses spatio-temporal information at different scales.}
\label{fig2}
\end{figure*}
    
    Generally, VideoQA includes the processes of feature extraction from each modality and the Video-Language (V-L) interaction for output. Many studies \cite{REF9,REF10,REF40,REF44} isolate these two processes, using fixed Convolutional Neural Network (CNN) \cite{REF1,REF11,REF20,REF53} for visual feature extraction and Recurrent Neural Network (RNN) \cite{REF12} or BERT \cite{REF14} for language embedding before the V-L interaction. We agree with \cite{REF15} that such suboptimal features acquired in an offline manner may not promise good fitting with the downstream multimodal tasks. For end-to-end VideoQA, more recent efforts \cite{REF19,REF45,REF47} are paid to pre-training transformers \cite{REF2} on large-scale datasets comprising vision-language pairs, e.g., COCO Captions \cite{REF16}, HowTo100M \cite{REF17}, and YT-Temporal-180M \cite{REF18}, and transferring knowledge to specific tasks. Despite their promising results, since powerful computational resources are still costly to access, we ask if via model engineering accurate end-to-end VideoQA can be achieved without using large-scale pre-training and huge feature extractors. We refer to such as \textit{\textbf{efficient}} end-to-end VideoQA. We would like to also note that, as we recognize the paradigm of learning from large amount of available data and generalizing to various downstream tasks as the promising route for end-to-end VideoQA, we believe this paper provides new supervised baselines and insights, informative for the future development of VideoQA and possibly V-L understanding.
    
    Targeting the above findings, we achieve efficient end-to-end VideoQA by building a more effective V-L interaction architecture. We deem the anisotropic pyramid structure as a promising candidate. The idea of pyramid is witnessed to help acquire multiscale spatial features for object detection \cite{REF48,REF49} and multiscale temporal features for dynamic scene classification \cite{REF51}, action recognition\cite{REF50}, and video grounding \cite{REF52} etc. Thereon, for the first time, we aim to incorporate the learning of multiscale spatial as well as temporal features, and leverage such spatio-temporal contexts to build the V-L interaction. As such, we establish multiscale, contextual, and spatio-temporal V-L interactions within a single model, essential for end-to-end VideoQA.
    
    While our evaluations on five VideoQA benchmarks demonstrate better or on-par performances against state-of-the-art methods, the computational complexity and cost of our method remain small and manageable. By simply leveraging feature extractors with reusable pre-trained weights loaded, our model is further improved and achieves competitive results for text-to-video retrieval. Our technical contributions are three-fold: (1) we propose a Pyramidal Multimodal Transformer (PMT) model for efficient end-to-end VideoQA, which works without using large-scale pre-training and huge feature extractors; (2) we propose a decomposition method within the pyramid to enable the learning of spatio-temporal features and their interactions with linguistic semantics at multiple scales; (3) we propose a contextual matching method to fuse spatio-temporal information of different scales, and a constraint function to maintain the integrity of local and global representations.

\section{Method}
    This section provides details of the video and text encoders, the V-L information passing and interactions within our PMT model, and the loss computations. By default, large-scale pre-training and huge feature extractors are not used. Additionally, the model is designed to be computational-efficient, with a comparably small number of trainable parameters. An overview of our PMT model is shown in Fig.\ref{fig2}.
    
\subsection{Video Encoder}
    Earlier studies usually use C3D \cite{REF11}, ResNet \cite{REF1,REF20}, and S3D \cite{REF53} for the spatio-temporal feature extraction from densely-sampled \cite{REF33,REF34} or multiscale-sampled video frames \cite{REF8}. Some also use Faster R-CNN \cite{REF38} to help acquire object-relevant features. While the recent end-to-end methods \cite{REF18,REF45,REF47} use ViT \cite{REF21}, TimeSformer \cite{REF54}, and hybrid ResNet/Vision Transformer as the video encoder, here we simply use X3D \cite{REF55} for its better computational efficiency.
    
    In short, we use the first five convolutional blocks of X3D as the video encoder, without using any down-sampling and pooling layers to maintain the rich spatio-temporal information of the grid-based feature map. Given video input $\mathcal{V}$, we acquire feature maps $\mathbf{X}=[\mathbf{x}_1,\mathbf{x}_2,...,\mathbf{x}_\mathrm{T}]\in\mathbb{R}^{\mathrm{T} \times \mathrm{H} \times \mathrm{W} \times \mathrm{C}}$, where $\mathrm{T},\mathrm{H},\mathrm{W},\mathrm{C}$ denote the number of frames, height, width, and number of channels of the group of feature maps. That is, for each video, we apply uniform sampling to acquire $\mathrm{T}$ frames. These feature maps are projected into the $D$-dimensional feature space using a $1\times1\times1$ 3D convolutional layer, with learnable positional embedding added to each feature matrix, thus we have $\mathbf{X}\in\mathbb{R}^{\mathrm{T} \times \mathrm{H} \times \mathrm{W} \times \mathrm{D}}$.
    
\subsection{Language Encoder}
    We use a trainable word embedding layer together with a single LSTM layer as the language encoder for text tokens and semantics extraction. In contrast to BERT-like models \cite{REF14} used in recent end-to-end V-L methods \cite{REF15, REF18, REF19, REF45, REF47}, a single LSTM layer is much efficient and proved to be effective for language encoding. 
    
    For each token of the language input, we first acquire the $D$-dimensional embedding with linear transformation layers. A bidirectional LSTM \cite{REF12} layer is further applied to acquire the contextual relations between tokens. The consequential language feature $\mathbf{G}\in\mathbb{R}^{\mathrm{T}_G \times \mathrm{D}}$ is obtained by concatenating the hidden states of LSTM in both directions per timestep, and $\mathrm{T}_G$ denotes the number of tokens.

\subsection{Pyramidal Video-Language Interaction}
    Apparently, the majority of end-to-end VideoQA research is focused on designing effective self- or semi-supervised tasks using datasets that comprise large amount of vision-language pairs, while improving the generalizability of a model to downstream tasks. Here, we demonstrate how to achieve efficient end-to-end VideoQA solely via a better design of V-L interactions within the proposed PMT model.
    
    \subsubsection{The Bottom-up Pathway.} While this work is inspired by the success of CNN \cite{REF1} on acquiring visual features at different levels, for VideoQA, we further consider how to acquire spatial and temporal features at different scales and enable their interaction with linguistic semantics. In the bottom-up pathway of PMT model, $\mathrm{L}$ blocks of multimodal transformer layers are stacked to acquire the V-L interaction at different spatio-temporal scales. For $l-$th block ($l\leq\mathrm{L}$), we introduce a \textit{\textbf{decomposition}} layer to divide the visual feature map $\mathbf{X}$ into its spatial part $\mathbf{S}^{(l)}$ and temporal part $\mathbf{M}^{(l)}$ as
    
\begin{figure}[t]
\centering
\includegraphics[width=0.8\columnwidth]{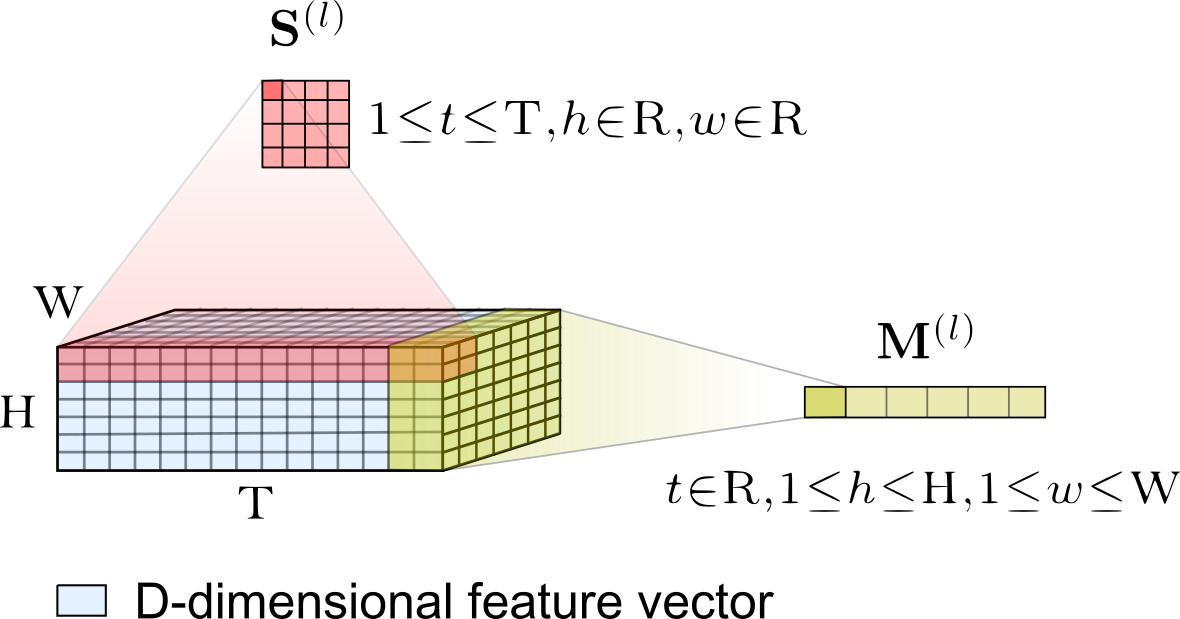}
\caption{The decomposition layer divides the visual feature into its spatial and temporal sub-features via dimension-wise max-pooling to help acquire multiscale spatio-temporal information within the pyramid.}
\label{fig3}
\end{figure}
    
\begin{equation}
    \mathbf{S}^{(l)}_{h^{'},w^{'},d}=\mathop{\max}\limits_{1\leq t\leq  \mathrm{T}, h\in \mathrm{R}, w\in \mathrm{R}}\mathbf{X}_{t,h,w,d},
\label{equa1}
\end{equation}
    
\begin{equation}
    \mathbf{M}^{(l)}_{t^{'},d}=\mathop{\max}\limits_{t\in \mathrm{R}, 1\leq h \leq  \mathrm{H}, 1\leq w \leq  \mathrm{W}}\mathbf{X}_{t,h,w,d},
\label{equa2}
\end{equation}
    where $\mathbf{X}_{t,h,w,d}$ denotes the digit at position $(t,h,w,d)$ of feature map $\mathbf{X}$, and $\mathrm{R}=2^{(l-1)}\times2^{(l-1)}$ denotes the size of non-overlapped spatial or temporal pooling region. The consequential $\mathbf{S}^{(l)}\in\mathbb{R}^{\frac{\mathrm{H}}{\mathrm{R}}\times \frac{\mathrm{W}}{\mathrm{R}}\times\mathrm{D}}$ contains complete global-temporal information and local-spatial details, whereas $\mathbf{M}^{(l)}\in\mathbb{R}^{\frac{\mathrm{T}}{\mathrm{R}}\times\mathrm{D}}$ contains complete global-spatial information and local-temporal details. An illustration of the decomposition is shown in Fig.\ref{fig3}. While such a decomposition method is based on max-pooling, per se, together with the pyramidal structure it efficiently reveals the multiscale property of the video in either spatial and temporal domains. By doing so, the computational cost is also reduced.
    
    For $l=1$, $\mathbf{S}^{(l)}$ and $\mathbf{M}^{(l)}$ are input to multimodal transformer block. For $l>1$, we use a residual merging to make the input for the block as
    
\begin{equation}
    \mathbf{S}^{(l)}=\mathbf{S}^{(l)}+\tilde{\mathbf{S}}^{(l-1)},
\label{equa3}
\end{equation}
\begin{equation}
    \mathbf{M}^{(l)}=\mathbf{M}^{(l)}+\tilde{\mathbf{M}}^{(l-1)},
\label{equa4}
\end{equation}
    where $\tilde{\mathbf{S}}^{(l-1)}$ and $\tilde{\mathbf{M}}^{(l-1)}$ are the multimodal-interacted output from the previous block. Within the $l-$th block, the V-L interaction of the transformer is implemented as
    
\begin{equation}
    \tilde{\mathbf{S}}^{(l)}={\mathop{FFN}}^{(l)}_\mathrm{S}(\mathop{LN}(\mathbf{S}^{(l)}_{att}))+\mathbf{S}^{(l)}_{att},
\label{equa5}
\end{equation}
\begin{equation}
    \tilde{\mathbf{M}}^{(l)}={\mathop{FFN}}^{(l)}_\mathrm{M}(\mathop{LN}(\mathbf{M}^{(l)}_{att}))+\mathbf{M}^{(l)}_{att},
\label{equa6}
\end{equation}
    with
\begin{equation}
\begin{split}
\mathbf{S}^{(l)}_{att}=\mathrm{concat}[{\mathop{MCA}}_h^{(l)}(\mathbf{S}^{(l)},\mathbf{S}^{(l)},\mathbf{G})]_{h=1}^\mathrm{H}\mathbf{W}_\mathrm{S}^{(l)}\\+\mathbf{S}^{(l)},
\end{split}
\label{equa7}
\end{equation}
\begin{equation}
\begin{split}
    \mathbf{M}^{(l)}_{att}=\mathrm{concat}[{\mathop{MCA}}_h^{(l)}(\mathbf{M}^{(l)},\mathbf{M}^{(l)},\mathbf{G})]_{h=1}^\mathrm{H}\mathbf{W}_\mathrm{M}^{(l)}\\+\mathbf{M}^{(l)},
\end{split}
\label{equa8}
\end{equation}
    where $\mathop{FFN}(\cdot)$ denotes the feed-forward network implemented with two ELU non-linear activation layers, $\mathop{LN(\cdot)}$ denotes layer normalization, $\mathrm{concat}[\cdot]$ denotes feature concatenation, and $\mathbf{W}_\mathrm{S}^{(l)}, \mathbf{W}_\mathrm{M}^{(l)}\in\mathbb{R}^{\mathrm{D}\times\mathrm{D}}$ are the trainable parameters. $\mathop{MCA(\cdot)}$ denotes multi-head cross-modal attention layer with $\mathrm{H}$ heads totally, which is implemented as

\begin{equation}
    {\mathop{MCA}}_h=\mathrm{softmax}(\frac{\mathbf{F}_q\mathbf{F}_k^\top}{\sqrt{\mathrm{D}}})\mathbf{F}_v,
\label{equa9}
\end{equation}
    where, for the input $(\mathbf{S}^{(l)},\mathbf{S}^{(l)},\mathbf{G})$ as an example, Query $\mathbf{F}_q=\mathop{LN}(\mathbf{S}^{(l)})\mathbf{W}_q^h$, Key $\mathbf{F}_k=\mathop{LN}(\mathbf{S}^{(l)})\mathbf{W}_k^h$, value $\mathbf{F}_v=\mathop{LN}(\mathbf{G})\mathbf{W}_v^h$, and $\mathbf{W}_q^h,\mathbf{W}_k^h,\mathbf{W}_v^h$ are the trainable parameters. Specially, for such an example of spatial-oriented information interaction, $\mathbf{S}^{(l)}$ is first flattened along the spatial dimensions before the computation of Equation \ref{equa9}.
    
    Unlike the isotropic transformer-based architecture, the stack of multimodal transformer blocks together with our decomposition method produces richer semantics information covering V-L interactions along the spatial and temporal dimensions at different scales. Such pyramidal structure also helps reduce computation loads.
    
    \subsubsection{The Top-down Pathway.} In order to acquire informative answering cues for VideoQA, the global semantics and local spatio-temporal details are both necessary. In our top-down pathway, for such an end, we match the information at different levels coming from both pathways. That is, each time the higher-level semantics with coarser spatio-temporal features are matched with lower-level semantics with higher spatio-temporal resolutions. Thereon, high-resolution as well as strong-semantics information are extracted. Here, we describe how to achieve this via lateral connections between the two pyramidal pathways and the proposed top-down Contextual Matching Block (CMB).
    
    For $l-$th level ($l<\mathrm{L}$) of the top-down pathway, the spatial output $\hat{\mathbf{S}}^{(l)}$ is computed as
    
\begin{equation}
    \hat{\mathbf{S}}^{(l)}={\mathop{CMB}}^{(l)}(\tilde{\mathbf{S}}^{(l)},\hat{\mathbf{S}}^{(l+1)},\hat{\mathbf{M}}^{(l+1)}),
\label{equa10}
\end{equation}
    with
\begin{equation}
    {\mathop{CMB}}_\mathrm{S}^{(l)}(\cdot)=\tilde{\mathbf{S}}^{(l)}+\mathbf{W}_{\mathrm{M}*\rightarrow\mathrm{S}}^{(l)}(\mathbf{W}_{\mathrm{S}\rightarrow\mathrm{M}*}^{(l)}\hat{\mathbf{S}}^{(l+1)}),
\label{equa11}
\end{equation}
\begin{equation}
    \mathbf{W}_{\mathrm{S}\rightarrow\mathrm{M}*}^{(l)}=\mathrm{softmax}(f(\hat{\mathbf{M}}^{(l+1)})f(\hat{\mathbf{S}}^{(l+1)})),
\label{equa12}
\end{equation}
\begin{equation}
    \mathbf{W}_{\mathrm{M}*\rightarrow\mathrm{S}}^{(l)}=\mathrm{softmax}(f(\tilde{\mathbf{S}}^{(l)})f(\hat{\mathbf{M}}^{(l+1)})),
\label{equa13}
\end{equation}
    where $f(\cdot)$ denotes the activated fully-connected layer. Similarly, for the temporal output $\hat{\mathbf{M}}^{(l)}$ we have
\begin{equation}
    \hat{\mathbf{M}}^{(l)}={\mathop{CMB}}^{(l)}(\tilde{\mathbf{M}}^{(l)},\hat{\mathbf{M}}^{(l+1)},\hat{\mathbf{S}}^{(l+1)}).
\label{equa14}
\end{equation}
    
    Specially, when $l=\mathrm{L}$, we have $\hat{\mathbf{S}}^{(l)}=\tilde{\mathbf{M}}^{(l)}$ and $\hat{\mathbf{M}}^{(l)}=\tilde{\mathbf{M}}^{(l)}$. In comparison with traditional top-down pathways that use up-sampling or direct attention-based match, the proposed CMB with cross-modal attention uses the spatial or temporal feature as the connection to extract relations between semantic features at different levels, while maintaining the integrity of spatio-temporal representations. Together with global-averaged language semantics $\bar{\mathbf{G}}$ that is acquired by averaging $\mathbf{G}$ across all the tokens, the output of PMT model at level $l$ is
    
\begin{equation}
    \mathbf{O}^{(l)}=\alpha\sum^{\frac{\mathrm{H}}{\mathrm{R}}\times\frac{\mathrm{W}}{\mathrm{R}}}_{t=1}\eta^{t}\hat{\mathbf{S}}^{(l)}+\beta\sum^{\frac{\mathrm{T}}{\mathrm{R}}}_{t=1}\gamma^{t}\hat{\mathbf{M}}^{(l)},
\label{equa15}
\end{equation}
    with
\begin{equation}
    \eta^{t}=\mathrm{softmax}(\bar{\mathbf{G}}\odot(\hat{\mathbf{S}}^{(l)})),
\label{equa16}
\end{equation}
\begin{equation}
    \gamma^{t}=\mathrm{softmax}(\bar{\mathbf{G}}\odot(\hat{\mathbf{M}}^{(l)})),
\label{equa17}
\end{equation}
    where $\odot$ denotes vector-wise inner product, $\eta^t$ and $\gamma^t$ denote the weights between language semantics and spatial or temporal feature, respectively. The learnable coefficients $\alpha$ and $\beta$ with $\alpha+\beta=1$ adaptively adjust the importance balance between spatial and temporal features when making the output given different tasks/samples. During inference, only the last output $\mathcal{O}^{(1)}$ is used for the task.
    
    We further introduce a constraint strategy using the multistep loss to help maintain the resolution and semantics integrity of features within this top-down pathway. For $\mathcal{L}^{(l)}$ computed from level $l$, the multistep loss is computed as
    
\begin{equation}
    \mathcal{L}_{step}=\sum^{\mathrm{L}-1}_{l=1}\mathrm{max}(0,\mathcal{L}^{(l)}-\mathcal{L}^{(l+1)}).
\label{equa18}
\end{equation}
    
    This multistep loss tunes the decoders to have smaller loss values along the descent levels. With a penalty factor $\lambda$, the total loss is
    
\begin{equation}
    %\mathcal{L}_{total}=\sum_{l=1}^{\mathrm{L}}\mathcal{L}^{(l)}+\mathcal{L}_{step}.
    \mathcal{L}_{total}=\mathcal{L}^{(1)}+\lambda\sum_{l=2}^{\mathrm{L}}\mathcal{L}^{(l)}+\mathcal{L}_{step}.
\label{equa19}
\end{equation}
    
    \subsection{Implementation Details}
    During our experiment, we use the PyTorch deep learning library and merely four NVIDIA GTX 1080 Ti GPUs. The video encoder, X3D-M, is initialized with Kinetics-400 \cite{REF56}. The GloVe \cite{REF22} embedding method is used to initialize the language encoder. In the top-right part of Figure \ref{fig2}, a two-layer fully-connected network with batch normalization is used as the decoder, while different loss functions are used to compute the loss $\mathcal{L}^{(l)}$ per decoder. For open-ended and repetition-count tasks in VideoQA, cross-entropy loss and mean square error are used respectively, and Hinge loss \cite{REF57} is used for the multi-choice task. For text-to-video retrieval conducted in our ablation study, similar to \cite{REF45}, we add contrastive loss to the video and language encoders and compute the binary cross-entropy loss given the last output $\mathcal{O}^{(1)}$.
    % It should be noted that, to train our PMT model, pre-training with video-language pairs is not used by default.
    
    For video processing, the number of frames is $\mathrm{T}=16$, the size of each frame is $\mathrm{H}=\mathrm{W}=224$. Simple yet efficient data augmentation \cite{REF20} is used to acquire more frames for training. That is, each input video is uniformly divided into $16$ segments and each segment contributes one frame, which is randomly cropped with width-height ratios of $0.8-1.2$, rotated, masked, and blurred. In testing, we directly use the frame in the middle of each segment and resize them to be $224\times224$. The feature dimension is $\mathrm{D=512}$. The number of heads is set to be $\mathrm{H}=8$ in the multimodal transformer block. The penalty factor $\lambda$ is set to 0.1. Adam optimizer is used with initial learning rate of $10^{-4}$, which is cut by half when the loss is not decreased for 10 epochs. The maximum number of epochs is 50, and the batch size is 32 for VideoQA, and 8 for text-to-video retrieval.

\begin{table*}[t]
\centering
\resizebox{\textwidth}{!}{
\begin{tabular}{l|lcc|cccc}
\hline
\textbf{Method} & \textbf{Video Rep.} & \textbf{Text Rep.} & \textbf{PT} & \textbf{Action}$\uparrow$  & \textbf{Trans.}$\uparrow$  & \textbf{FrameQA}$\uparrow$ & \textbf{Count}$\downarrow$\\ 
\hline
ST-TP \cite{REF10}  &ResNet, C3D &Glove &- & 62.9  & 69.4  & 49.5 & 4.32      \\
Co-Mem \cite{REF27} &ResNet-152, Flow CNN &Glove &- & 68.2  & 74.3  & 51.5 & 4.10 \\
PSAC \cite{REF23}   &ResNet-152 &Glove &- & 70.4  & 76.9  & 55.7 & 4.27 \\
STA \cite{REF67}         &ResNet-152 &Glove &- & 72.3  & 79.0  & 56.6 & 4.25 \\
%HME \cite{REF26} &ResNet, C3D &Glove &- & 73.9  & 77.8  & 53.8 & 4.02\\
L-GCN \cite{REF30}  &ResNet-152, Mask R-CNN &Glove &- & 74.3  & 81.1  & 56.3 & 3.95 \\
QueST \cite{REF25}  &ResNet-152 &Glove &- & 75.9  & 81.0  & 59.7 & 4.19 \\
HCRN \cite{REF33}   &ResNet, ResNeXt-101 &Glove &- & 75.0  & 81.4  & 55.9 & 3.82 \\
B2A \cite{REF34}    &ResNet, ResNeXt-101 &Glove &- & 75.9  & 82.6  & 57.5 & 3.71 \\
HAIR \cite{REF36}   &ResNet, Faster R-CNN &Glove &- & 77.8  & 82.3  & 60.2 & 3.88\\
PVI-Net \cite{REF41}&ResNet, C3D &BERT &- & 79.2  & 84.7  & 60.3 & 3.79 \\
MASN \cite{REF35}   &I3D, Faster R-CNN &Glove &- & 84.4  & 87.4 & 59.5 & 3.75   \\
HQGA \cite{REF43}   & ResNeXt-101, Faster R-CNN &BERT &- & 76.9 & 85.6 & 61.3 & - \\
MHN \cite{REF8}         & ResNet, 3D ResNet152 &Glove &-     & 83.5 & 90.8 & 58.1 & 3.58 \\
\hline
ClipBERT \cite{REF15} &ResNet-50 &BERT &COCO,VG  & {(82.8)} & {(87.8)}  & {(60.3)}  & -  \\
SiaSamRea \cite{REF19}&ResNet-50 &BERT &COCO,VG & {(79.7)} &  {(85.3)} &  {(60.2)} &{(3.61)}   \\
HD-VILA \cite{REF47}  &ResNet, TimeSformer &BERT & HD-VILA-100M  & {(84.3)}  & {(90.0)} & {(60.5)} & -   \\
\hline
\textbf{PMT (ours)}   &X3D-M &Glove &-  & \textbf{87.6} & \textbf{92.9} & 60.6 & \textbf{3.41}   \\ 
\hline
\end{tabular}
}
\caption{Comparison with state-of-the-art methods on TGIF-QA datasets. Video Rep. and Text Rep. denote the video and text encoders, respectively. PT denotes the datasets used for pre-training. Results in brackets denote what acquired with large-scale pre-training. Only our PMT model and methods placed in the second section are end-to-end.}
\label{table1}
\end{table*}

\section{Experiments}
    We first report the comparison results of our method against a series of state-of-the-art methods on five VideoQA benchmarks, with more insights provided in the ablation study.

\subsection{Datasets}
    We use the state-of-the-art benchmarks for VideoQA in our experiment: 1) TGIF-QA \cite{REF10}, a large-scale VideoQA dataset comprising 72K animated GIFs and 165K question-answer pairs, which are divided into four task types, namely multi-choice tasks of Action and Transition (Trans.), open-ended task of FrameQA, and Count that requires answering the exact integer number; 2) MSVD-QA \cite{REF9}, with 1,970 short clips and 50,505 open-ended question-answer pairs; 3) MSRVTT-QA \cite{REF9,REF58}, with 10K videos and 243K question-answer that have the same setting of MSVD-QA; 4) ActivityNet-QA \cite{REF4}, with 5.8K complex videos (average duration of 3 mins) downloaded from the internet and 58K question-answer pairs; and 5) Youtube2Text-QA \cite{REF70}, where the video data come from MSVD-QA with 9.9K question-answer pairs, which are divided into three task types, namely \textit{what, who, other}, we experiment with the multi-choice task. We use the official split of training, validation, and testing sets provided by the datasets, and report results acquired on the testing set. 
    
    For each videoQA dataset, we pre-define a vocabulary comprising the top $K$ frequent words appeared in the training set, and $K$ is set to 4000 for MSVD-QA and 8000 for the rest. We report accuracy for open-ended and multi-choice tasks. To compute the mean squared error for Count task in TGIF-QA dataset, we apply a rounding-function to the model output to acquire the predicted integer result. For the text-to-video retrieval experiment, following \cite{REF6,REF15,REF17,REF72}, we use the 7K training set and report results on the 1K testing set. Therein, the first $k$ (R@$k$) and median rank (MdR) accuracies are reported.
    
\begin{table}[t]
\centering
\resizebox{\linewidth}{!}{
\begin{tabular}{l|ccc}
\hline
\textbf{Method} & \textbf{MSVD-QA}$\uparrow$ & \textbf{MSRVTT-QA}$\uparrow$ & \textbf{ActivityNet-QA}$\uparrow$  \\ 
\hline
AMU (2017)    & 32.0  & 32.5  & -        \\
E-SA (2019)   & 27.6   & 29.3  & 31.8     \\
HME (2019)    & 33.7  & 33.0   & -    \\
%TSN ()   & 36.7     & 35.4  & -    \\
HGA (2020)    & 34.7  & 35.5   & -     \\
%QueST \cite{REF25}    & 36.1  & 34.6   & -      \\
HCRN (2020)   & 36.1  & 35.6   & -    \\
B2A (2021)    & 37.2  & 36.9   & -    \\
HAIR (2021)   & 37.5  & 36.9  & -     \\
SSML (2021)   & {(35.1)}  & {(35.1)}  & -     \\
CoMVT (2021)  & 35.7 {(42.6)} & 37.3 {(39.5)}   & 36.6 {(38.8)}         \\
VQA-T (2021)  & 41.2 {(46.3)}   & 39.6 {(41.5)}      & 36.8 {(38.9)}       \\
MHN (2022)    & 40.4 & 38.6   & -     \\
HQGA (2022)   & 41.2 & 38.6   & -     \\
\hline
ClipBERT (2021)    & -              &  {(37.4)}       & -                  \\
SiaSamRea (2021)   & {(45.5)}       &  {(41.6)}       &  {(39.8)}    \\
ALPRO (2022)       & 41.5 {(45.9)}  & 39.6 {(42.1)}   & -                   \\
HD-VILA (2022)     & -              & {(40.0)}        & -              \\
\hline
\textbf{PMT (ours)} & 41.8 & 40.3 & \textbf{42.6}      \\ 
\hline
\end{tabular}
}
\caption{Comparison with state-of-the-art methods on MSVD-QA, MSRVTT-QA, and ActivityNet-QA datasets. Results in brackets denote what acquired with large-scale pre-training. Only our PMT model and methods placed in the second section are end-to-end.}
\label{table2}
\end{table}
    
\setcounter{table}{2}
\begin{table}[t]
\centering
\resizebox{0.85\linewidth}{!}{
\begin{tabular}{l|lcccc}
\hline
\textbf{Method} &\textbf{What}$\uparrow$ &\textbf{Who}$\uparrow$ &\textbf{Other}$\uparrow$  &\textbf{All}$\uparrow$    \\ 
\hline
r-ANL (2017)  &63.3 &36.4 &84.5 & 52.0    \\
HME (2019)    &83.1 &77.8 &86.6 & 80.8    \\
L-GCN (2020)  &86.0 &81.5 &80.6 & 83.9    \\
HAIR (2021)   &87.8 &82.4 &81.4 & 85.3    \\
% \hline
\textbf{PMT (ours)}        &\textbf{90.8} &80.4 &\textbf{99.0} & \textbf{86.4}   \\
\hline
\end{tabular}
}
\caption{Comparison with state-of-the-art methods on Youtube2Text-QA dataset. Only PMT model is end-to-end.}
\label{table3}
\end{table}
    
\setcounter{table}{3}
\begin{table*}[t]
\centering
\resizebox{0.7\linewidth}{!}{%
\begin{tabular}{l|c|c|cccc} 
\hline
\multicolumn{1}{c|}{\multirow{2}{*}{\textbf{Method}}} & \multirow{2}{*}{\textbf{PT}} & \textbf{VideoQA} & \multicolumn{4}{c}{\textbf{Text-to-video retrieval}} \\
\multicolumn{1}{c|}{} &  & acc $\uparrow$ & R1$\uparrow$ & R5$\uparrow$ & R10$\uparrow$ & MdR$\downarrow$ \\ 
\hline
SSML (2021) & HowTo100M & 35.1 & 17.4 & 41.6 & 53.6  & 8 \\
VQA-T (2021) & HowToVQA69M & 41.5 & - & - & - & - \\
ATP (2022) & WebImageText & - & 27.8 & 49.8 & - & - \\
\hline
HERO (2020) & HowTo100M & - & 16.8 & 43.4 & 57.7 & - \\
ClipBERT (2021) & COCO,VG & 37.4 & 22.0 & 46.8 & 59.9 & 6 \\
SiaSamRea (2021) & COCO,VG & 41.6 & - & - & - & - \\
VideoCLIP (2021) & HowTo100M & - & 30.9 & 55.4 & 66.8  & - \\
FiT* (2021) & WebVid2M, CC3M & - & 31.0 & 59.5 & 70.5 & 3 \\
ALPRO (2022) & WebVid2M, CC3M & 42.1 & 33.9 & 60.7 & 73.2 & 3 \\
\hline
\textbf{PMT (ours)} & Weights from WebVid2M, CC3M & \textbf{42.4} & 31.0 & 55.5 & 66.2 & 4 \\
\hline
\end{tabular}
}
\caption{The experiment on scalability of our PMT model using MSRVTT dataset. PT denotes the datasets used for pre-training. Only our PMT model and methods placed in the second section are end-to-end.}
\label{table4}
\end{table*}
    
\subsection{Comparison with State-of-the-arts}
    Table \ref{table1} reports the results on TGIF-QA dataset. Without using large-scale pre-training (e.g., using COCO Captions \cite{REF16}, Visual Genome Captions \cite{REF60}, HD-VILA-100M video-text pairs \cite{REF47}) and huge or complex feature extractors (e.g., C3D \cite{REF11}, I3D \cite{REF63}, ResNeXt \cite{REF20}, Faster R-CNN \cite{REF38}, and even a tool for text analysis \cite{REF59} seen in B2A \cite{REF34}), our efficient end-to-end PMT model achieves the best performances in Action ($+3.2$), Trans. ($+2.1$), and Count ($-0.17$) tasks of TGIF-QA dataset. We also find that HQGA, ClipBERT, and HD-VILA ignored the Count task of this dataset. It is possible that their methods do not work well on searching global and local semantics in data of long temporal durations.
    
    Table \ref{table2} reports the results on MSVD-QA, MSRVTT-QA, and ActivityNet-QA datasets. For comparison, extra methods are added, namely AMU \cite{REF9}, E-SA \cite{REF4,REF9}, HME \cite{REF26}, HGA \cite{REF29} SSML \cite{REF39}, CoMVT \cite{REF40}, VQA-T \cite{REF44}, and ALPRO \cite{REF45}. It should be mentioned that, the offline methods CoMVT and VQA-T are pre-trained on HowTo100M \cite{REF17} and HowToVQA69M \cite{REF44}, respectively. For ALPRO, WebVid2M \cite{REF61} and CC3M \cite{REF62} are used in pre-training. When pre-training is not used in these methods, our method achieves the best performances across the three datasets. We also recognize that pre-training indeed largely improves the performances of several methods on these three datasets. However, for the ActivityNet-QA dataset, our method outperforms the methods that are pre-trained on large-scale datasets.
    
    % We would like to mention again that, pre-training with HowTo100M \cite{}, HowToVQA69M \cite{}, WebVid2M \cite{}, and CC3M \cite{} that seen in methods we compare here is not practical in our experimental environment.
    
    %In table \ref{table3}, we draw another comparison on Youtube2Text-QA dataset with another method r-ANL \cite{REF70} added, where we also achieve improved performances when the three attributes (\textit{what, who,} and \textit{other}) are pooled together on multi-choice ($+1.1$) and open-ended ($+0.5$) tasks. Here, the compared methods L-GCN and HAIR all require the use of extra object detectors.
    In table \ref{table3}, we draw another comparison on Youtube2Text-QA dataset with another method r-ANL \cite{REF70} added, where we also achieve improved performances when the three attributes (\textit{what, who,} and \textit{other}) are pooled together on multi-choice ($+1.1$) tasks. Here, L-GCN and HAIR all require the use of extra object detectors.
    
    In general, while large-scale pre-training contribute to VideoQA particularly in MSVD-QA and MSRVTT-QA datasets, the better or on-par performances achieved by our method across the five VideoQA benchmarks so far shall demonstrate the equal importance of proposing effective V-L interactions in a model for VideoQA. This is important at this moment when expensive computational resources are not easily accessible.
    
\subsection{Comparison on Computational Efficiency}
\begin{figure}[t]
\centering
\includegraphics[width=\columnwidth]{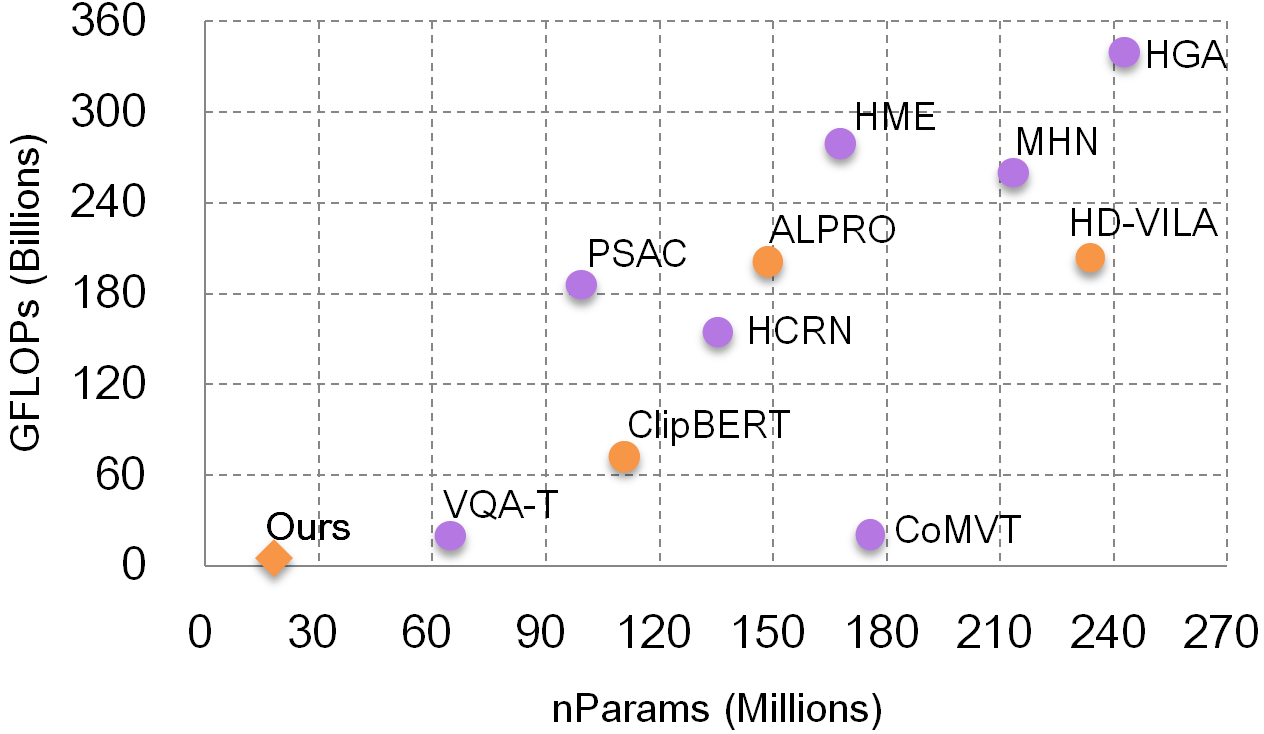}
\caption{Comparison on computational efficiency. The end-to-end methods are marked in yellow, and those using offline feature extractors are marked in purple.}
\label{fig4}
\end{figure}

    We propose efficient end-to-end VideoQA in this work with our PMT model, in comparison with methods that normally adopt large-scale pre-training and huge feature extractors. We compute the number of trainable parameters (nParams) and GFLOPs \cite{REF1,REF55,REF53} of our model and several recent methods, which are shown in Figure \ref{fig4}. We set the number of input frames as 16 for computing GFLOPs. It should be noted that parameters and computational loads created by object detectors used in some of these methods are left out. Our PMT model (nParams=18.1M, GFLOPs=5.01B) is more efficient than other end-to-end methods, e.g., ClipBERT (nParams=110.7M, GFLOPs=72.2B), ALPRO (nParams=148.5M, GFLOPs=201.1B) and HD-VILA (nParams=233.6M, GFLOPs=203.6B). Additionally, except for being suboptimal, methods that adopt pretrained and \textit{frozen} encoders may have smaller compute need in training.

\subsection{Ablation Study}

    \subsubsection{The Scalability of our PMT Model.} One way to alleviate the dependency on expensive computational resources and take the advantage of large-scale pre-training is to import the feature extractors that have pre-trained weights loaded. Here, we conduct an experiment with our PMT model, where we use TimeSformer and BERT that used in \cite{REF45} to replace the vanilla feature extractors used in our experiments above. These two feature extractors were pre-trained on WebVid2M and CC3M video-text pairs by the authors of \cite{REF45}. We experiment with both the VideoQA and text-to-video retrieval tasks of MSRVTT dataset \cite{REF9,REF58}. 
    
    As reported in Table \ref{table4}, by taking the advantage of feature extractors that have more powerful representation learning capacities, our method outperforms other methods on VideoQA, and achieves competitive if not better performances on text-to-video retrieval. Specifically, we acquire an $+0.3$ improvement on VideoQA, and outperforms SSML, ATP \cite{REF46}, HERO \cite{REF72}, ClipBERT, and VideoCLIP \cite{REF71}. For FiT \cite{REF61}, * denotes the results acquired by training on the 9K training set. It should be noted that, despite the feature extractors, our PMT model is not pre-trained on large-scale datasets.
    
    Aside from the promising scalability of our method suggested in this experiment, the generalizability of our method for different V-L tasks is demonstrated. This experiment also implies that model engineering (as conducted in this work) together with the reusable prior knowledge (weights from a one-shot pre-training) may enable the end-to-end V-L understanding in an efficient and environmental-friendly manner.
    
    \subsubsection{The Impact of the Proposed Pyramidal Structure.} This paper proposes to enable multiscale, contextual, and spatio-temporal V-L interactions within a single pyramidal network. Here, we use the default PMT model adopted in the experiments above as the baseline method, and examine the performance of its several candidate variants on TGIF-QA dataset. We remove the decomposition method used in the bottom-up pathway (w/o decomp.), thus features become isotropic. For others, we replace the proposed top-down contextual matching block with up-sampling (w/ up-sample), or attention-based fusion (w/ attention), which are used in previous pyramidal networks, and remove the multistep loss (w/o constraint) to show its importance on maintaining the feature integrity. As shown in Table \ref{table5}, performances drop for all the tasks when features become isotropic in the network, since the learning of multiscale information becomes difficult. The use of up-sampling and attention-based fusion in the top-down pathway also leads to worse results on all the tasks, showing the importance of our proposed contextual matching block. Without the constraint created by the multistep loss on feature integrity across local and global semantics, performances decrease for all the tasks.

\setcounter{table}{4}
\begin{table}[t]
\centering
\resizebox{\linewidth}{!}{
\begin{tabular}{lcccc}
\hline
\textbf{Method}  & \textbf{Action}$\uparrow$  & \textbf{Trans.}$\uparrow$  & \textbf{FrameQA}$\uparrow$ & \textbf{Count}$\downarrow$\\ 
\hline
PMT w/o decomp.   & 85.4          & 90.1          & 58.6          & 3.50 \\
PMT w/ up-sample  & 86.7          & 92.3          & 59.5          & 3.46 \\
PMT w/ attention  & 86.5          & 92.4         & 60.1          & 3.46 \\
PMT w/o constraint   & 87.5          & 92.6          & 60.4          & 3.45 \\
PMT (default)     & \textbf{87.6} & \textbf{92.9} & \textbf{60.6} & \textbf{3.41}  \\
\hline
\end{tabular}
}
\caption{Ablation study on our pyramidal}
\label{table5}
\end{table}

    \subsubsection{The Impact of Tunable Hyperparameters.} 
\begin{figure}[t]
\centering
\includegraphics[width=\linewidth]{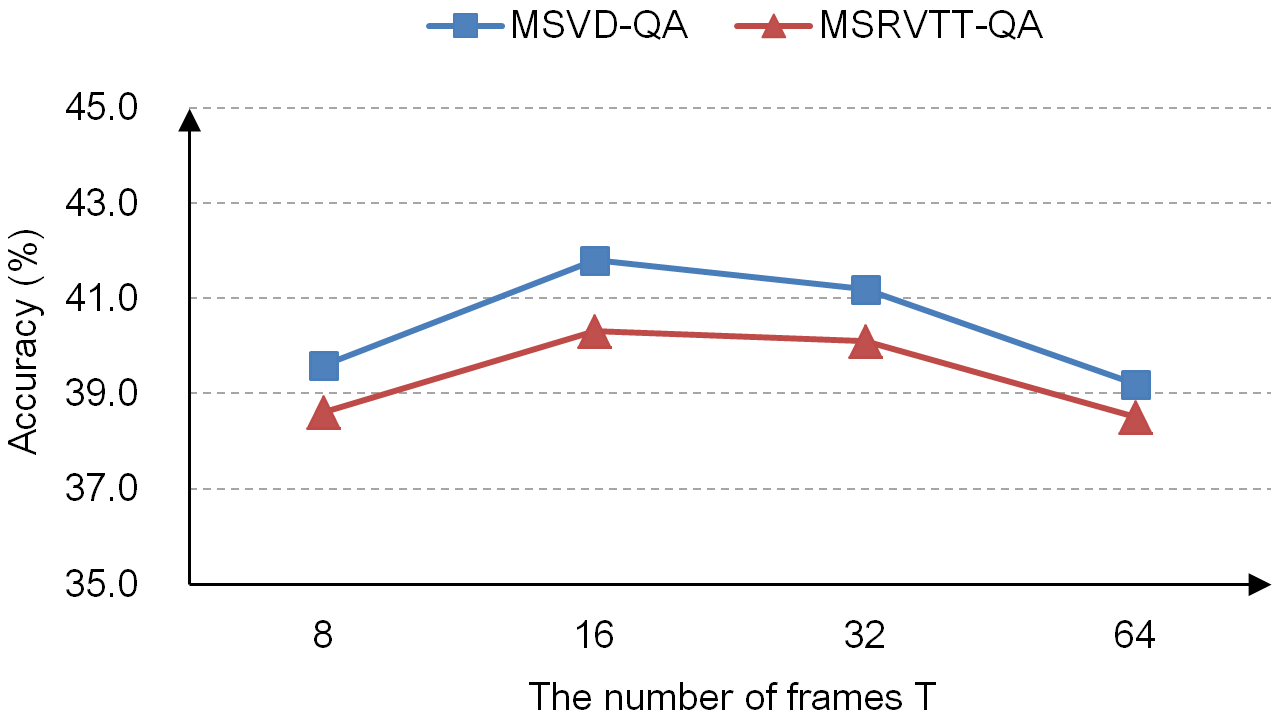}
\caption{The impact of number of input frames.}
\label{fig5}
\end{figure}

One of the noticeable hyperparameter is  the number of input frames that balances the amount of information provided to the model and the computational load or even the introduction of irrelevant noise. Here, we experiment with $\mathrm{T}=\{8,16,32,64\}$, using MSVD-QA and MSRVTT-QA datasets, while default values of other hyperparameters are still used. As shown in Figure \ref{fig5}, our PMT model reaches the best performance at $\mathrm{T}=16$, and the change of such a hyperparameter leads to obvious performance fluctuations. Another hyperparameter that could be of interest is the number of levels $\mathrm{L}$ of our pyramidal architecture that sets another balance between informative representation learning and computational loads or ever the risk of overfitting. Particularly, the maximum value of $\mathrm{L}$ is set by $\log_2(\mathrm{T})$, given the decomposition method used in our model. We conduct another experiment on MSVD-QA dataset, and do not find further improvements on performance for different values other than $\mathrm{L}=3$ when $\mathrm{T}=16$.

\section{Conclusion}
    This paper proposes a new Pyramidal Multimodal Transformer (PMT) model for efficient end-to-end Video Question Answering (VideoQA). Particularly, we enable multiscale, contextual, and spatio-temporal V-L interactions within this single model. Without using large-scale pre-training and huge or complex feature extractors, our PMT model achieves better or on-par performances against state-of-the-art approaches across five VideoQA benchmarks. We demonstrate the scalability and generalizability of our method on text-to-video retrieval, by leveraging feature extractors that have reusable pre-trained weights loaded. Our work suggests that, aside from the booming of large-scale pre-training, model engineering is equally important to V-L understanding, especially when powerful computational resources are still expensive to access and environmental-friendly research is encouraged. Our future work will consider the evaluation with datasets like Next-QA \cite{REF73} and AGQA \cite{REF74}, which are certainly more demanding on spatio-temporal and commonsense reasoning.

\section{Acknowledgments}
    This work is funded by the National Natural Science Foundation of China (62106247).

\small
\bibliography{aaai23}

\end{document}